# A Hassle-Free Machine Learning Method for Cohort Selection of Clinical Trials

Liu Man

liumanlalala@gmail.com

Traditional text classification techniques in clinical domain have heavily relied on the manually extracted textual cues. This paper proposes a generally supervised machine learning method that is almost equally hassle-free and does not use clinical knowledge. The employed methods were simple to implement, fast to run and yet effective.

This paper proposes a novel named entity recognition (NER) based ensemble system capable of learning the keyword features in the document. Instead of merely considering the whole sentence/ paragraph for analysis, the NER based keyword feature extraction system can stress the important clinic relevant phase more. In addition, to capture the semantic information in the documents, the FastText (Grave, 2017) features originating from the document level FastText classification results are exploited. The architecture of our proposed method is shown in Figure 1. For clarity, our model is described in two parts: feature extractor and ensemble classification Model.

**Feature Extractor**

**Key word Feature Extraction** The keyword features are extracted using NER-based method and traditional TFIDF method. This paper proposes using clinic relevant keywords extracted with the open source clinic NER model cliNER[1] (Boag, 2015) as additional features. The predicted named entities tags (i.e. "treatment", "problem", "test") are also incorporated into the feature vectors. Adding clinic relevant keywords and context-sense identities to the feature vectors will increase the performance of the classification. To avoid the deficiency of information in the clinical words, the traditional TFIDF features that help capture more information in the document are also adopted.

**FastText Feature Extraction** Besides the keyword features, we employ the FastText features to obtain the document level feature. The FastText features are acquired by importing FastText classification probabilities in the feature vectors. The open source

---

[1] https://github.com/text-machine-lab/CliNER

FastText classification model[2] is employed. As the experiments suggest, adding FastText features can improve the final result.

**Gazetteer Feature Extraction** Additionally, a small hand-crafted gazetteers are utilized to excerpt the class-unique information in the text. If the word in the document fits the entries in the gazetteer, the context of this word will serve as the gazetteer features. The context size and weights vary with the class label. Four gazetteers are employed, and each of which merely contains the most relevant words. The largest gazetteer is diet supplement list with size of 120. The average size of gazetteers is 51.

**Context Feature Extraction** Similar to the gazetteer features, we extract context features for some class labels (i.e. class "CREATININE", class "ASP-FOR-MI", class "HBA1C", class "KETO-1YR". For instance, if the word "creatinine" occurs in the document, the context words should be extracted and added into the feature vectors. The context size and weight depend on the class labels. Experiments reveal that context features of different classes can benefit the classification result of the other class. For example, the context features of class "ADVANCED-CAD" and "CREATININE" can ameliorate the prediction result of class "ASP-FOR-MI".

**Ensemble Classification System**

To solve the biasness caused by using one machine learning classification method, we propose an ensemble system that combines three components, i.e. a logistic regression (LR) model, a linear SVM model and a gradient boosting decision tree (GBDT) model (Ye, 2009). The noted components are assembled by predicting the class label based on the argmax of the sums of the predicted probabilities. Before ensemble, the components are experimented respectively to establish the weights in the ensemble system. Given the case of a multi-class classification problem occurring in this task, 13 classifiers are built for the 13 class labels.

We perform extensive 5-fold cross validation experiments to confirm the weight of each type of features on different class labels and the weight of each classification model in the ensemble system. The final official result (micro F1) of our team is obtained as 83.00%.

---

[2] https://github.com/facebookresearch/fastText

Figures:

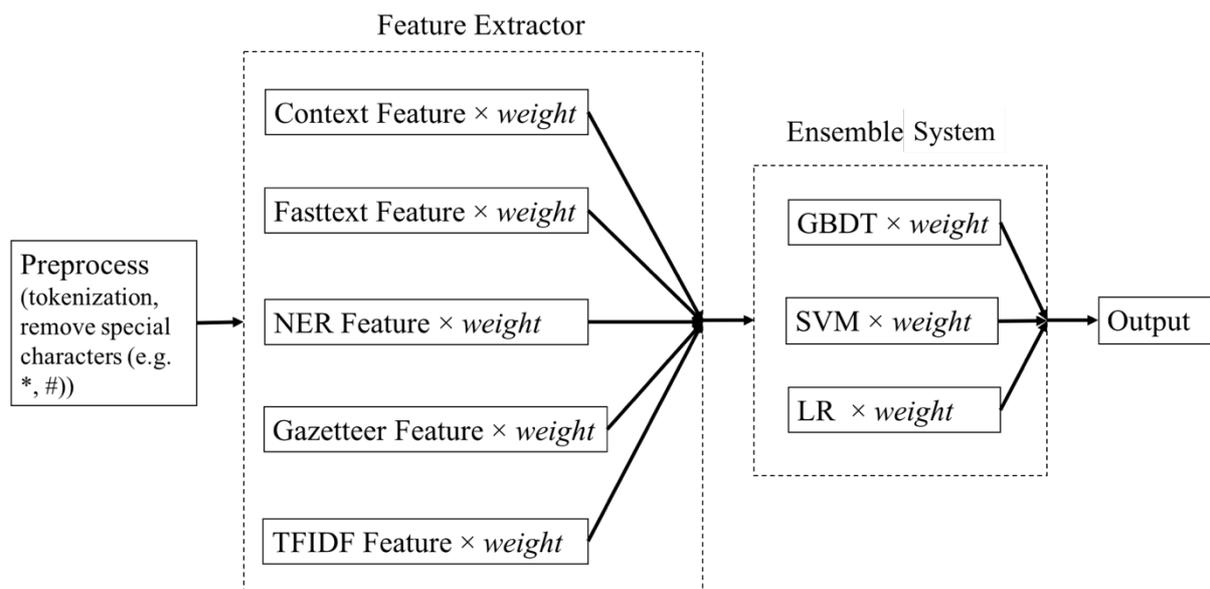

Figure 1: The general architecture of the proposed approach. GBDT represents the gradient boosting decision tree model and LR represents the logistic regression model. All the weights are tuned by 5-fold cross validation experiments.